# Optical Remote Sensing Image Understanding with Weak Supervision:
# Concepts, Methods, and Perspectives

Jun Yue, Leyuan Fang, *Senior Member, IEEE,* Pedram Ghamisi, *Senior Member, IEEE,* Weiying Xie, *Member, IEEE,* Jun Li, *Fellow, IEEE,* Jocelyn Chanussot, *Fellow, IEEE,* and Antonio Plaza, *Fellow, IEEE*

*Abstract*—In recent years, supervised learning has been widely used in various tasks of optical remote sensing image understanding, including remote sensing image classification, pixel-wise segmentation, change detection, and object detection. The methods based on supervised learning need a large amount of high-quality training data and their performance highly depends on the quality of the labels. However, in practical remote sensing applications, it is often expensive and time-consuming to obtain large-scale data sets with high-quality labels, which leads to a lack of sufficient supervised information. In some cases, only coarse-grained labels can be obtained, resulting in the lack of exact supervision. In addition, the supervised information obtained manually may be wrong, resulting in a lack of accurate supervision. Therefore, remote sensing image understanding often faces the problems of incomplete, inexact, and inaccurate supervised information, which will affect the breadth and depth of remote sensing applications. In order to solve the above-mentioned problems, researchers have explored various tasks in remote sensing image understanding under weak supervision. This paper summarizes the research progress of weakly supervised learning in the field of remote sensing, including three typical weakly supervised paradigms: 1) Incomplete supervision, where only a subset of training data is labeled; 2) Inexact supervision, where only coarse-grained labels of training data are given; 3) Inaccurate supervision, where the labels given are not always true on the ground.

This work was supported in part by the National Natural Science Foundation of China under Grant 61922029 and Grant 62101072, in part by the Hunan Provincial Natural Science Foundation of China under Grant 2021JJ30003 and Grant 2021JJ40570, in part by the Science and Technology Plan Project Fund of Hunan Province under Grant 2019RS2016, in part by the Key Research and Development Program of Hunan under Grant 2021SK2039, and in part by the Scientific Research Foundation of the Hunan Education Department under Grant 20B022 and Grant 20B157. *(Corresponding author: Leyuan Fang.)*

Jun Yue is with the Department of Geomatics Engineering, Changsha University of Science and Technology, Changsha 410114, China (e-mail: jyue@pku.edu.cn).

Leyuan Fang is with the College of Electrical and Information Engineering, Hunan University, Changsha 410082, China, and also with the Peng Cheng Laboratory, Shenzhen 518000, China (e-mail: fangleyuan@gmail.com).

Pedram Ghamisi is with the Helmholtz-Zentrum Dresden-Rossendorf (HZDR), Helmholtz Institute Freiberg for Resource Technology, 09599 Freiberg, Germany, and also with the Institute of Advanced Research in Artificial Intelligence (IARAI), 1030 Vienna, Austria (e-mail: p.ghamisi@gmail.com).

Weiying Xie is with the State Key Laboratory of Integrated Services Networks, Xidian University, Xi'an 710071, China (e-mail: wyxie@xidian.edu.cn).

Jun Li is with the Guangdong Provincial Key Laboratory of Urbanization and Geo-simulation, School of Geography and Planning, Sun Yat-sen University, Guangzhou 510275, China (e-mail: lijun48@mail.sysu.edu.cn).

Jocelyn Chanussot is with Université Grenoble Alpes, Inria, CNRS, Grenoble INP, LJK, 38000 Grenoble, France (e-mail: jocelyn.chanussot@grenoble-inp.fr).

Antonio Plaza is with the Hyperspectral Computing Laboratory, Department of Technology of Computers and Communications, University of Extremadura, 10003 Cáceres, Spain (e-mail: aplaza@unex.es).

## I. Introduction

MACHINE learning has played a very important role in the development of optical remote sensing image (RSI) understanding, especially the methods that are based on supervised learning [1], [2]. To a large extent, this is due to the rise and development of deep learning, which has a strong ability to extract abstract features without requiring the manual design of features [3], [4], [5], [6], [7]. With the increasing availability of open-source machine learning frameworks such as TensorFlow [8] and PyTorch [9] and the continuous iteration of a large number of deep learning methods, high-quality RSI understanding models are being widely introduced and promoted the depth and breadth of remote sensing applications in the field of geoscience. In the supervised machine learning paradigm, prediction models are usually learned from training datasets containing a large number of high-quality training samples [10], [11], [12], [13].

To achieve good performance using supervised RSI understanding methods, some preconditions need to be met, including: 1) Each sample needs to have a label; 2) All labels need to be fine-grained, and 3) All labels need to be free of noise. However, in practical applications, many conditions are difficult to fully meet [14], [15]. In order to better meet the needs of actual RSI understanding, researchers began to explore weakly supervised learning methods, including incomplete supervised methods (only a subset of training samples have labels), inexact supervised methods (some training samples have no fine-grained labels) and inaccurate supervised methods (some training samples have wrong labels) [16], [17], [18], [19]. Fig. 1 shows the three weakly supervised paradigms.

Incomplete supervision refers to a situation where a small number of labeled samples can be obtained, but it is insufficient to train a high-weight model, and a large number of unlabeled samples is available [20]. Formally, the strong supervision paradigm is to train a model from the training dataset $\{(x_1, y_1), ..., (x_m, y_m)\}$, where there are $m$ labeled training samples in the training dataset. Meanwhile, a typical incomplete supervision paradigm is to train a model from the training dataset $\{(x_1,y_1),...,(x_n,y_n),x_{n+1},...,x_m\}$, where there are $n$ labeled training samples and $u = m - n$ unlabeled samples. The research of incomplete supervision mainly focuses on active learning and semi-supervised learning [16]. In the field of RSI understanding, the typical applications are hyperspectral image (HSI) pixel-wise classification based on active learning [21], [22], [23], [24], [25] and RSI scene classification based on semi-supervised learning [26], [27], [28], [29].

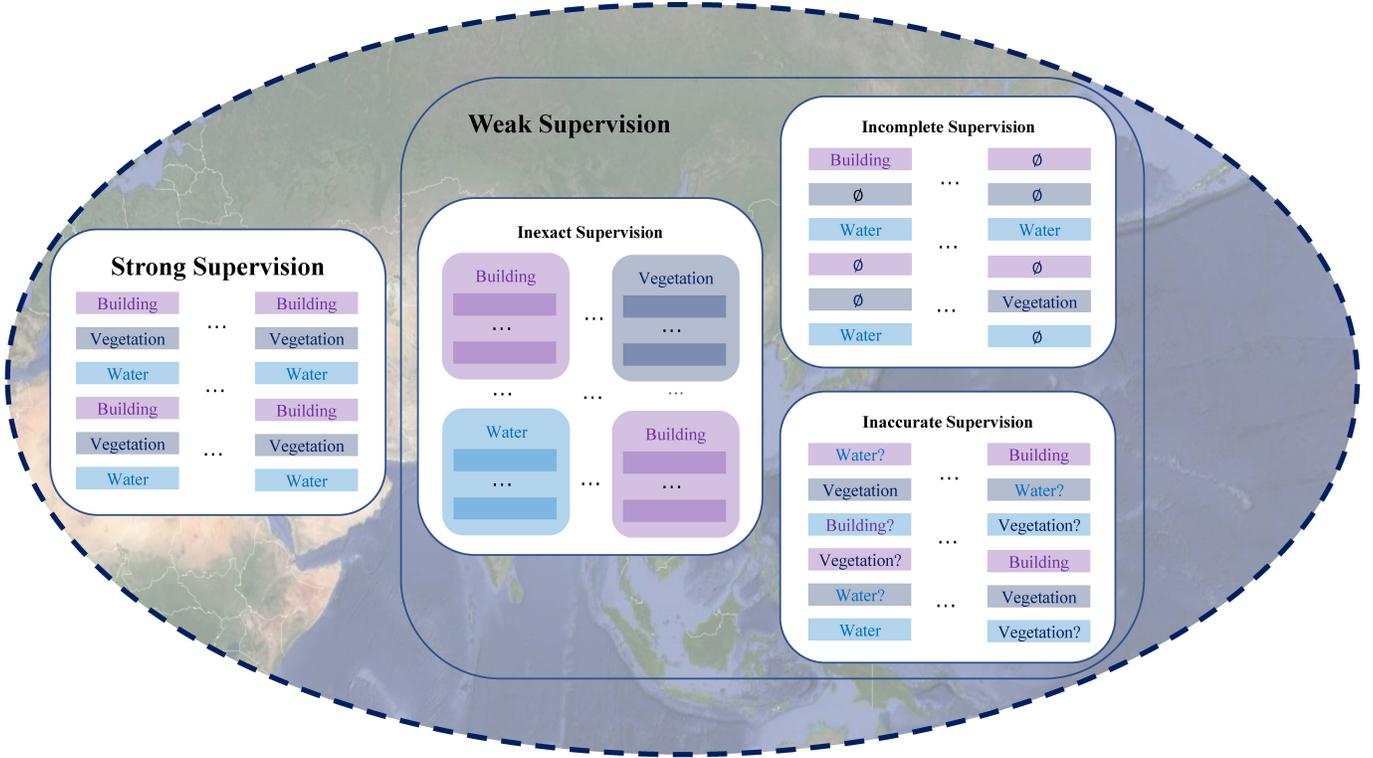

Fig. 1: Illustration of three weakly supervised paradigms, i.e., incomplete supervision, inexact supervision and inaccurate supervision. A classification task is taken as an example here, in which the categories of the ground objects are vegetation (represented by a gray rectangle), water (represented by a blue rectangle) and building (represented by a purple rectangle), respectively.

Inexact supervision refers to a situation where supervised information is given during training, but the supervised information is not as exact as expected [17], [30]. Formally, a typical inexact supervision paradigm is to train a model from the training data set $\{(X_1,y_1),...,(X_i,y_i),...\}$, where $X_i = \{x_1,...x_{m_i}\}$ is a collection of training samples (i.e., bag), $m_i$ is the total number of training samples in the bag. In this paradigm, there is at least one sample in the bag $X_i$ whose label is $y_i$. The research of inexact supervision is mainly focused on multi-instance learning. In the field of RSI understanding, the typical applications are Very-High-Resolution (VHR) remote sensing object localization and detection based on multi-instance learning [31], [32], [33].

Inaccurate supervision refers to a situation that the supervised information is not always true. In other words, the labels of some samples may be affected by noises and errors [34], [35], [36]. Formally, a typical inaccurate supervi-sion paradigm is to train a model from the training dataset $\{(x_1,y_1),...,(x_i,y_i),...\}$, where the label $y_i$ may contain errors. The research of inaccurate supervision mainly focuses on how to suppress the influence of noisy labels. In the field of RSI understanding, the typical applications are HSI pixel-wise classification with noisy labels [37], [38], [39], VHR scene classification with noisy labels [40], [41], [42] and crowdsourcing in RSI understanding [43], [44], [45], [46].

In the field of RSI understanding, more and more researchers are exploring methods based on weakly supervised learning to overcome the difficulty of obtaining supervised information. At the same time, various weak supervision paradigms have also been used and improved the performance of RSI understanding. In this paper, optical RSI refers to the image of the Earth's surface detected by visible and infrared sensors. Beyond the scope of conventional weakly supervised research [47], [48], [49], weakly supervised learning for RSI understanding usually needs to consider how to make full use of spatial and spectral features. Therefore, this paper is dedicated to sorting out RSI understanding methods based on weakly supervised learning, and forming a clear framework for RSI understanding with weak supervision. The purpose of this paper is to enable researchers to more accurately locate their own research in the overall research landscape with weak supervision. In addition, this paper also found some gaps to be filled between weakly supervised learning and RSI understanding, providing some research ideas for future research.

The structure of this paper is as follows. In the second part, incomplete supervision and its typical applications in RSI understanding, including active learning for HSI classification and semi-supervised learning for VHR RSI scene classification, are summarized in detail. In the third part, inexact supervision and its typical applications in RSI understanding, including multi-instance learning for RSI object localization and detection, are summarized in detail. In the fourth part, inaccurate supervision and its typical applications in RSI understanding, including HSI classification and VHR scene classification with noisy labels and crowdsourcing for RSI understanding, are summarized in detail. In the fifth part, the application of weakly supervised learning in RSI understanding is summarized and the future directions are defined.

## II. OPTICAL REMOTE SENSING IMAGE UNDERSTANDING WITH INCOMPLETE SUPERVISION

Incomplete supervision involves an RSI understanding task, that is, only a small amount of ground labeled data is obtained during model training, which is not enough to train a suitable model effectively [50], [51]. In remote sensing, however, there are usually a large number of unlabeled data available, which can be fully used to assist model training. At present, there are mainly two kinds of incomplete supervision methods in RSI understanding: active learning with human intervention [52], [53], [54], [55] and semi-supervised learning without human intervention.

Active learning attempts to obtain the unlabeled samples that are most helpful to improve the accuracy, and submits them to human experts for labeling. Through the intervention of human experts, the ground-truth of selected unlabeled instances can be obtained [56], [57], [58]. In contrast, semi-supervised learning improves the learning performance by exploring the data distribution and automatically using unlabeled data other than labeled data without human intervention [59], [60], [61].

### A. RSI Understanding with Active learning

Active learning assumes that the labels of unlabeled samples can be obtained by human intervention [16], [62]. It is generally assumed that the labeling cost depends on the number of samples labeled by human experts. Therefore, one of the tasks of active learning is to minimize the number of samples submitted to human experts under the condition of ensuring learning performance, so as to minimize the cost of training a good model. In other words, active learning attempts to select the most valuable unlabeled samples and submit them to human experts. In order to minimize the labeling cost, given some labeled samples and a large number of unlabeled samples, active learning will select the unlabeled samples to maximize the final performance. There are two effective criteria for unlabeled sample selection, namely, informativeness criterion and representativeness criterion [52], [63].

Informativeness criterion is used to measure the extent to which unlabeled instances help to reduce the uncertainty of statistical models [64], [65], while representativeness criterion is used to measure the extent to which instances help to represent the structure of its corresponding class [55], [66], [16], [67]. Typical methods based on informativeness criterion are Query-By-Committee (QBC) and Uncertainty Sampling (US). The QBC method generates multiple models to form a committee, in which each member represents a model with parameters. Each model selects the unlabeled sample with the most labeling value. Finally, the unlabeled samples selected the most times are the unlabeled samples that need to be labeled [68], [69]. The US method selects the instances with the highest uncertainty and send them to human experts [70]. Representativeness criterion-based methods usually adopt clustering-based methods to select unlabeled samples [71], [72].

Active learning has achieved great success in supervised RSI understanding because it can select training samples with the highest discrimination [73], [74], [75], [76]. As a sampling method with bias, active learning tends to select samples in low-density areas. However, ground object classes in HSIs usually have inter-class correlation [77], [78], [79], [80]. The emergence of this problem limits the potential of active learning to select valuable unlabeled samples. To solve this problem, the concept of feature-driven active learning is introduced [81], in which, sample selection is carried out in a given optimized feature space. The results reported in [81] revealed that the method improves the potential of active learning in HSI classification. In addition, in order to make full use of the spectral information and spatial contextual information of HSI in sample selection, an active learning method based on a Bayesian network has been proposed. Experimental results on three real HSI datasets show the effectiveness of this method [24].

### B. Semi-supervised learning for RSI Understanding

Semi-supervised learning attempts to utilize unlabeled data without involving human experts [82], [83], [84], [85]. Generally speaking, semi-supervised learning can also be divided into transductive learning and pure semi-supervised learning [86], [87]. The main difference between transductive learning and pure semi-supervised learning lies in the corresponding testing data [88], [89], [90]. Transductive learning assumes that unlabeled samples are considered to be the test data, that is, the purpose of learning is to obtain the best accuracy on these unlabeled samples [91], [92], [93]. In pure semi-supervised learning, the test data is not given in advance, that is, the unlabeled data given is not used for model testing, so the over-fitting problem needs to be seriously considered [94], [95]. Fig. 2 illustrates the difference between active learning, transductive learning and pure semi-supervised learning [16], [17].

One of the interesting questions about semi-supervised learning that may arise is why unlabeled data can help us to learn better RSI understanding models. Fig. 3 illustrates the role of unlabeled samples in helping train better models. For a classification task, as shown in Fig. 3 (its categories are *building*, *water*, and *vegetation*), if there are only three labeled samples and one unlabeled sample, we can only rely on a random guess to classify the unlabeled sample because the unlabeled data point is just in the middle of the three labeled data points. However, if more unlabeled data points are available, we can predict that the object class of the unlabeled data point is *building* with high confidence. In this example, although the ground truth labels of these unlabeled data points are not given, researchers can explore the implicit data distribution information through semi-supervised learning, which can improve the performance of the model [96].

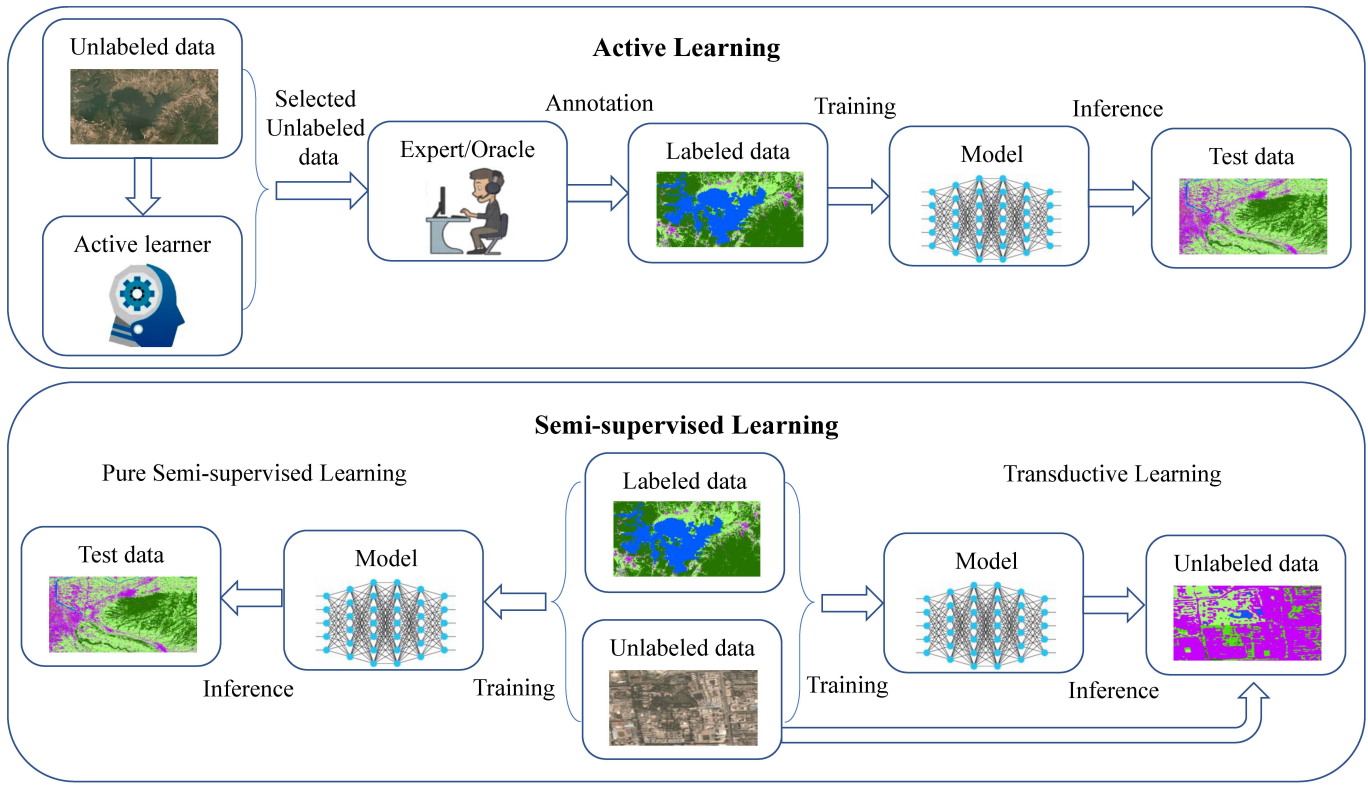

Fig. 2: Illustration of three incomplete supervised paradigms, i.e., active learning, transductive learning and pure semi-supervised learning.

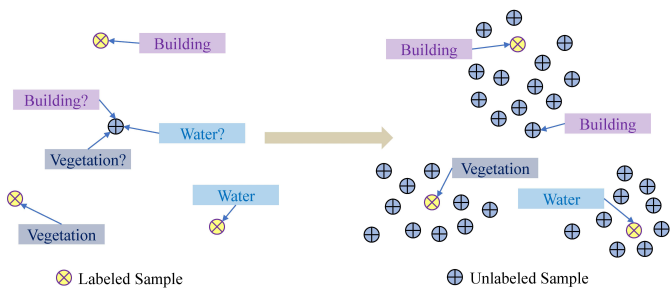

Fig. 3: Illustration of the effectiveness of unlabeled samples on ground object classification. Semi-supervised learning helps to explore the implicit data distribution information, which helps to improve the performance.

As for the data distribution in semi-supervised learning, there are clustering hypothesis and manifold hypothesis in the machine learning theory community [59]. The clustering hypothesis assumes that samples with the same class are in the same cluster, that is, samples belonging to the same cluster set can be classified into the same class. Under this assumption, a large number of unlabeled samples can be used to help explore the dense and sparse regions in the feature space, so as to guide the semi-supervised learning algorithm to adjust the decision boundary, make it pass through the sparse region in the feature space, and prevent separating the samples within the same cluster [97], [98].

The manifold hypothesis assumes that the sample data we observed can be regarded as the expression of low dimensional manifold in high-dimensional space. Compared with clustering hypothesis, manifold hypothesis mainly considers the local structures of samples. Under this assumption, a large number of unlabeled examples can be used to fill the feature space, which helps to describe the characteristics of local regions more accurately, so that the model can better fit the data [99]. There are three kinds of semi-supervised learning methods for RSI understanding: regularization-based methods [100], [101], [102], [103], generation-based methods [26], [104], [105], [106], and graph-based methods [107], [108], [109], [110], [111].

In VHR RSI classification, since a single visual feature can only describe one aspect of the ground object, it is usually necessary to generate multiple features of ground objects and concatenate them to obtain better classification results. In order to make full use of unlabeled samples in this process, a multi-graph fusion framework based on semi-supervised manifold learning (SSM-MF) has been proposed [114]. This method combines multiple features to learn a low dimensional feature space, which can effectively describe the semantic information of both the labeled and unlabeled data. The main advantages of this method are as follows: 1) The geometric information in labeled data and the structural information in unlabeled data are fully utilized; 2) The complementarity of multiple features is explored to effectively prevent the impact of the curse of dimensionality. Experiments on remote sensing data sets show the effectiveness of this method [114].

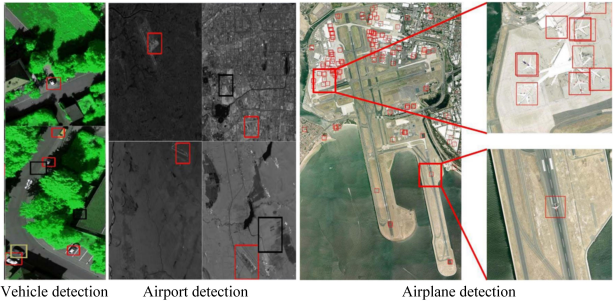

Fig. 4: An application example of inexact supervision for object localization and detection in RSI analysis. The example comes from [47], [112], [113].

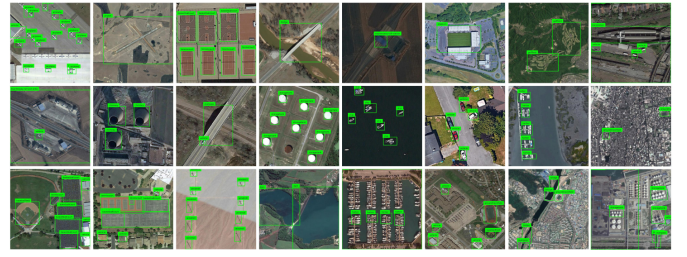

Fig. 5: Illustration of the RSI object detection results of PICR on the DIOR dataset under inexact supervision. The example comes from [122].

VHR RSI scene classification is a scene-level classification task [29]. Because of its wide application, it is urgent to improve its accuracy. However, due to the difficulty of obtaining high-quality VHR RSI labels, the performance of scene classification is difficult to be further improved. To solve this problem, a semi-supervised generative framework (SSGF) has been proposed. This method includes deep feature extraction module, self-labeling module, and discrimination evaluation module to realize unlabeled data information extraction and scene classification. The experimental results on three real datasets show that SSGF can extract valuable features from unlabeled samples [26].

## III. OPTICAL REMOTE SENSING IMAGE UNDERSTANDING WITH INEXACT SUPERVISION

Inexact supervision involves the situation that some supervised information is given, but it does not exactly match with strong supervised information [17]. The typical scenarios in the understanding of RSIs are as follows: 1) Object localization with image-level labels of a given RSI; 2) Object detection with image-level labels of a given RSI [115], [116], [117], as shown in Fig. 4.

Because the method of remote sensing object localization and detection in the scenario of inexact supervision mainly uses manual annotation at the image-level, the learning framework not only needs to solve the typical problems with strong supervision, such as the change of appearance and scale within the same class, and bounding box regression, but also needs to solve the challenges caused by the inconsistency between human annotation and the actual ground situation [118], [119]. In remote sensing object localization and detection with inexact supervision, the accuracy of bounding box regression is closely related to the process of model learning. The key is to obtain the learnable bounding box-level supervised information based on the given image-level information. In this process, the bounding box information obtained has great uncertainty. Therefore, in this kind of weakly supervised learning paradigm, it is inevitable that there will be a lot of noisy and ambiguous information when the weak supervision is propagated [31], [120], [47].

In order to effectively improve the precision and recall of ground object localization and class recognition in inexact supervision scenario, the existing methods are usually divided into two stages: initialization stage and refinement stage. In the initialization stage, the image-level annotation is propagated to the bounding box annotation based on the prior knowledge, so that the noisy, biased and low-quality supervision information can be generated for further bounding box regression. In the refinement stage, the effective features of ground objects are usually learned based on the bounding box annotation generated in the initialization stage, and finally qualified object localization and detection models can be obtained for RSI understanding [116]. Researchers can make improvements to generate bounding box annotation with more accurate location and more accurate labels, which is of great help to improve the performance of the model in terms of mean Average Precision (mAP) in the initialization stage. However, due to the limitations of the annotation quality generated in the initialization stage, in the refinement stage, researchers can improve the robustness of the learning method to meet the challenges of inaccurate, biased, and insufficient bounding box annotation [47]. By effectively improving the performance of each stage, an acceptable weak supervised object locator and detector can be trained [106], [121].

Learning with inexact supervision for RSI object localization and detection has attracted much attention due to its wide range of applications. It only needs to provide image-level annotation, which greatly reduces the cost of application. At present, in the process of propagating image-level annotation to bounding box annotation, most existing methods select a ground object instance with the highest score from a large number of initial object proposals to train object locator and detector. However, in large-scale and complex RSIs, there are usually multiple object instances belonging to the same class. Therefore, selecting only one object instance with the highest score to train the model will lose a lot of important information. Such methods may highlight the most represen-tative part of the RSI and ignore the details. To solve this problem, an end-to-end progressive context instance refinement (PCIR) method has been proposed [122]. This method mainly consists of two strategies: the Dual Context Instance Refinement (DCIR) strategy and the Progressive Proposal Self-Pruning (PPSP) strategy. Specifically, the DCIR strategy aims to transfer the focus of the detection network from local parts to the entire image and further to several potential proposals by using local and global contextual information. The PPSP strategy reduces the influence of complex background in RSIs by dynamically rejecting negative instance proposals. Comprehensive experiments on two well-known datasets show that this method can significantly improve the detection accuracy compared with the existing methods. Fig. 5 shows the results of PCIR on the object DetectIon in Optical RSIs (DIOR) dataset under inexact supervision [122].

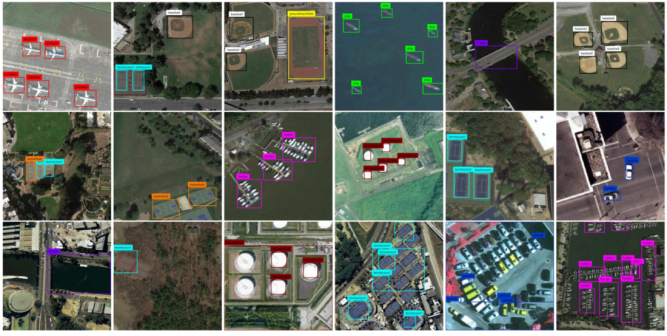

Fig. 6: Illustration of the RSI object detection results of TCANet under inexact supervision. The example comes from [123].

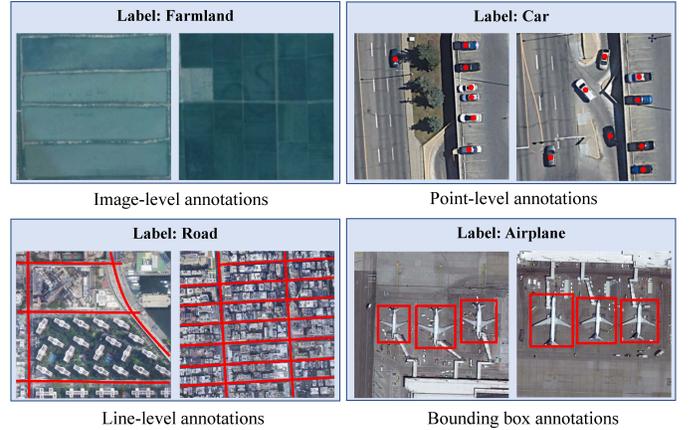

Fig. 7: Illustration of four levels of inexact supervised annotations, i.e., image-level annotations, point-level annotations, line-level annotations, and bounding box annotations.

Another challenge in propagating image-level annotations to bounding box annotations is that many instances of the same class often appear in adjacent locations. In this case, the existing methods usually take the adjacent overlapping instances as the same proposal. In order to solve this problem, a Triple Context-Aware Network (TCANet) has been proposed to explore the contextual information of different regions in RSI, so as to distinguish the characteristics of different regions [123]. This method is mainly composed of two modules: the Global Context Awareness Enhancement (GCAE) module and the Dual Local Context Residual (DLCR) module. Specifically, the GCAE module activates the features of the whole object by capturing the global contextual information. The DLCR module captures instance-level discrimination information by using the semantic discrepancy of the local region [123]. Fig. 6 shows the results of TCANet under inexact supervision.

In RSI understanding, other common inexact supervised annotations are point-level annotations and line-level annotations, and these four levels of annotations (i.e., image-level annotations, point-level annotations, line-level annotations, and bounding box annotations) are shown in Fig. 7. Based on point-level annotations of RSIs, researchers proposed an RSI object detection method to minimize the cost of labeling and improve the detection performance [124]. In this research, point-level annotations are introduced to guide the generating of the candidate proposals and the pseudo bounding boxes. Then, the detection model is trained by using the pseudo boundary boxes. This method includes a progressive candidate bounding box mining strategy to improve the accuracy of detection. The experimental results on a VHR RSI dataset show that the algorithm has better performance than the You Only Look Once (YOLO) v5 algorithm [124]. Based on line-level annotations of RSIs, researchers proposed a road label propagation method for road surface segmentation [125]. In addition, researchers have proposed several learning methods combining multiple inexact supervised annotations, which can also be applied to RSI understanding in the future [126], [127].

## IV. Optical Remote Sensing image understanding with Inaccurate Supervision

Inaccurate supervision refers to situations where the supervised information of RSI does not fully reflect the real situation on the ground, and the label information of some samples may be wrong [128]. In the field of RSI understanding, a typical scenario is to train an RSI understanding model with good performance in the presence of noise in the training set [129]. In many existing theoretical studies on training with noisy labels, most of them assume that the distribution of noise is random, that is, the noisy labels are affected by random noise. In order to eliminate the influence of noisy labels, a traditional method is to identify the labels polluted by noise and correct them. Recently, a scenario of RSI understanding with inaccurate supervision is crowdsourcing, which is a label collection method that distributes the labeling of samples to volunteers. Since volunteers are not necessarily professionals, it is generally believed that their labeling results may be inaccurate, that is, model learning with the labeled data generated by crowdsourcing is a kind of weakly supervised learning [130], [16]. Researchers try to find a method that can not only improve the labeling efficiency, but also basically ensure the labeling accuracy [131].

### A. RSI Understanding with Noisy Labels

Samples with noisy labels usually appear in data sets automatically collected from the Internet, and are mislabeled by non-expert annotators or even experts in challenging tasks (such as crater classification in RSIs [132], [133], [134]). Although the model based on deep learning has made great progress in various fields of RSI understanding and significantly improved the accuracy of the existing models, a problem to be solved is that the noisy samples in the training dataset will reduce the generalization ability of the model, so that the model may be overfitted. The model based on deep learning relies on a large number of correctly labeled samples, but with the continuous growth of remote sensing data sets, it is very difficult to ensure the correctness of each label. Therefore, it is very important to consider the existence of labeling noise during the training of RSI understanding models [135], [136].

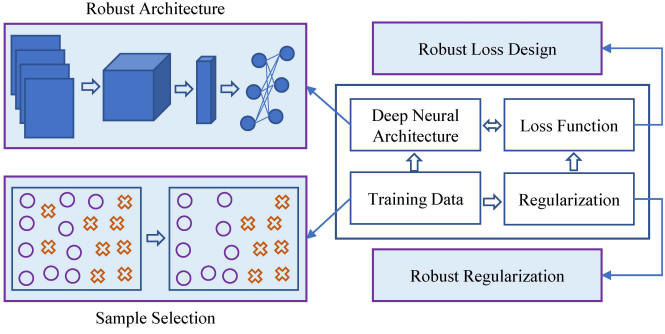

Fig. 8: Four types of methods to improve the robustness of learning with noise, i.e., robust architecture, robust regularization, robust loss design, and sample selection [135].

Fig. 9: Different types of noise transition matrix. (a) Symmetric noise transition matrix. (b) Asymmetric noise transition matrix.

One of the methods to improve the accuracy of RSI understanding is to train the deep model with large-scale correctly labeled training samples. Unfortunately, labeling a large number of RSIs is very expensive and time-consuming. Therefore, researchers can use more economical alternative methods, such as crowdsourcing [45], [137] and online query [138]. These methods can save costs by organizing a large number of non-professionals to label samples. But these methods will inevitably bring noise to the training data set. At the same time, noisy samples may also appear in small-scale data sets when the task of sample labeling is very difficult, or the opinions of the labeling personnel are different. In the case of manually labeling RSIs, it is difficult to avoid noisy labels in the training data set [139]. At present, the main sources of noisy labels can be classified into four types: 1) The lack of contextual information of ground objects in low-resolution RSIs leads to the low confidence of labels; 2) Errors caused by the negligence of labeling personnel; 3) Ambiguity caused by multi-source labeling; 4) Ambiguity caused by data encoding [136], [140].

In this paper, noisy samples refer to the samples whose labels are different from their real ground object classes. When the noisy sample is mentioned in this paper, it does not mean that there is noise in the input RSI, but there is noise in the label of the sample. From the existing literature, the methods of learning with noise are mainly divided into the following categories: robust architecture, robust regularization, robust loss design, and sample selection, as shown in Fig. 8 [135]. Formally, typical learning with noisy labels paradigm is to train a model from the training dataset $\{(x_1, y_1), ..., (x_i, y_i), ...\}$, where the true label and the observed label of $x_i$ are denoted as $y_i^\dagger$ and $y_i$, respectively. We use $\lambda \in [0, 1]$ to denote the overall noise rate. $\vartheta_{ij}$ is the probability that object class $j$ is wrongly divided into object class $i$, i.e., $\vartheta_{ij} = P(y_i = i | y_i^\dagger = j)$. In the current literature on learning with noise, there are different kinds of noise in the training data set:

1) Symmetric noise: symmetric noise is also known as random noise or uniform noise, which means that labels have the same probability of being misclassified into another ground object class [141], [142], as shown in Fig. 9 (a).

2) Asymmetric noise: asymmetric noise means that for different ground object classes, their probabilities of being misclassified into another ground object class are not completely consistent [143], as shown in Fig. 9 (b).

3) Open-set noise: noisy labeling problems can be divided into two types: closed-set noise problem and open-set noise problem. The problem of closed-set noise occurs when all real labels belong to known classes. Open-set noise refers to a situation when the sample has a wrong label that is not included in the known training data set [144], [145], [136], [146].

Most of the initially proposed methods for RSI understanding with noisy labels are based on estimating the noise transition matrix to understand the mechanism that how the correct labels of the RSI are wrongly divided into other ground object classes [147], [148], [149], as shown in Fig. 9. The loss function for model learning with noise transition matrix can be defined as follows:

$$\begin{cases} \mathcal{L}^\theta = \dfrac{1}{N} \sum_{i=1}^{N} -log P(y = y_i | x_i, \theta) \\ P(y = y_i | x_i, \theta) = \sum_{c=1}^{N_c} P(y = y_i | y^\dagger = c) P(y^\dagger = c | x_i, \theta) \end{cases} \quad (1)$$

where $N$ and $N_c$ are the total number of training samples and the total number of object classes, respectively [150], [136], [151].

A typical RSI understanding with noisy labels method based on the noise transition matrix of different noise types is the random label propagation algorithm (RLPA). The core idea of this method is to mine knowledge from the observed HSIs (such as spectral-spatial constraints based on super-pixels) and apply it to the label propagation process. Specifically, RLPA first constructs a spectral-spatial probability transformation matrix (SSPTM) which considers both spectral similarity and spatial information based on super-pixels. Then some training samples are randomly selected as initial labeled samples, and the remaining samples are set as unlabeled samples. SSPTM is used to propagate the label information from the initial labeled samples to the remaining unlabeled samples. Multiple labels can be obtained for each training sample by repeated random sampling and propagation. The labels generated by label propagation can be determined by a majority voting algorithm [152].

Another kind of RSI understanding with noisy labels is the loss correction method. Such methods usually add regularization term, weighting term or attention term to the loss function to reduce the low confidence prediction that may be related to noisy samples. The advantage of these methods is that they can be applied to the existing methods without modifying the network structure, but only modifying the loss function [153], [154]. A typical method is robust normalized soft maximum loss (RNSL). This method uses the negative Box-Cox transformation to replace the logarithmic function of the normalized softmax loss (NSL) to reduce the influence of noisy samples on the learning of corresponding prototypes, thereby improving the robustness of NSL [155]. In addition, a truncated robust normalized softmax loss (t-RNSL) through threshold-based truncation of loss has been proposed, which can further enhance the generation of object prototypes based on the HSI features with high similarity, so that intra-class objects can be clustered well and inter-class objects can be separated well. Experiments on two benchmark datasets demonstrate the effectiveness of this method in three tasks (i.e., classification, retrieval, and clustering) [155].

In order to reduce the influence of noisy samples, researchers have proposed noisy sample selection and elimination strategies. The typical one is super-pixel to pixel weighting distance (SPWD) [156]. This method first detects the noisy samples and tries to remove the influence of the noisy label in the training set on the model training. This method uses the spectral and spatial information in HSI based on the following two assumptions: 1) The corresponding ground object classes of the pixels in a super-pixel are consistent; 2) The adjacent pixels in the spectral space correspond to the same ground object class. The method comprises the following steps. First, the adaptive spatial information of each labeled sample is generated based on super-pixel segmentation. Second, the spectral distance information between each super-pixel and pixel is measured. Third, the decision function based on density threshold is used to remove the noisy label in the original training set. The classification accuracy of the SPWD detection method is evaluated by using support vector machine (SVM) classifier. Experiments on several actual HSI data sets show that this method can effectively improve the HSI classification accuracy in the presence of noisy labels [156].

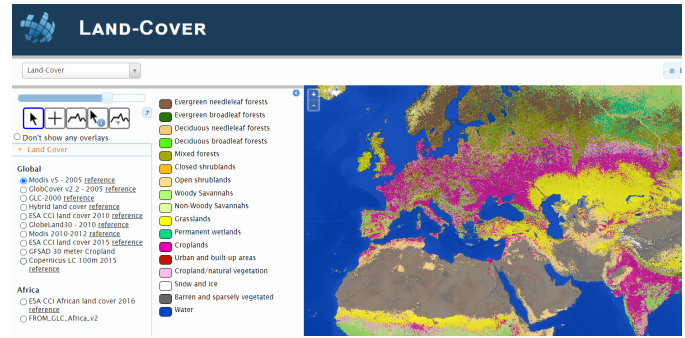

Fig. 10: The Geo-Wiki platform [162], which enables volunteers from all over the world to provide rich supervised information to help produce accurate global land cover maps, etc.

## B. RSI Understanding with Crowdsourcing

Crowdsourcing refers to the use of the group's ability to collect data to solve specific problems. Through crowdsourcing, problems that are difficult for a single person can be easily solved, and the time and cost required to solve the problem are greatly reduced. The success of crowdsourcing is based on the fact that group efforts usually produce better results than individual efforts at a similar time [157]. Through crowdsourcing, a very large-scale RSI understanding task can be decomposed into many sub-tasks, and volunteers are required to complete these small, and easy sub-tasks. In this way, it is possible to complete a large amount of tedious labeling work in a short period of time. Since crowdsourcing can be used in many different areas, various terms have been proposed to express the same idea [158], including volunteered geographic information [159] and citizen science [160]. Although different from the contents embodied in these terms, their workflow is basically the same, which can be used to collect supervised information for RSI understanding tasks [157], [161].

Geo-Wiki is a typical crowdsourcing project for RSI understanding. It is a platform created by the International Institute for Applied Systems Analysis (IIASA) in 2009 that can make full use of the public's energy and wisdom to label remote sensing data. Through public participation, it helps to verify the existing geographic information and collect new labels of the ground objects by using tools such as google satellite image and Bing map. Volunteers can provide valuable field data about the types of ground objects by visual interpretation, comparing the existing data with satellite images or collecting new data for labeling. The data collected by volunteers can be input through traditional desktop platforms or mobile devices (such as Android phones) [163]. In addition to the traditional incentive methods, Geo-Wiki also encourages volunteers to provide information by holding activities and games. Geo-Wiki has a large number of registered volunteers and many successful crowdsourcing projects, collecting a variety of data for various tasks, including global building data, global vegetation cover data, global shrub cover data, global snow cover data, global wetland cover data, and global farmland cover data [164]. Fig. 10 shows the land classification map created on the Geo-Wiki platform [137].

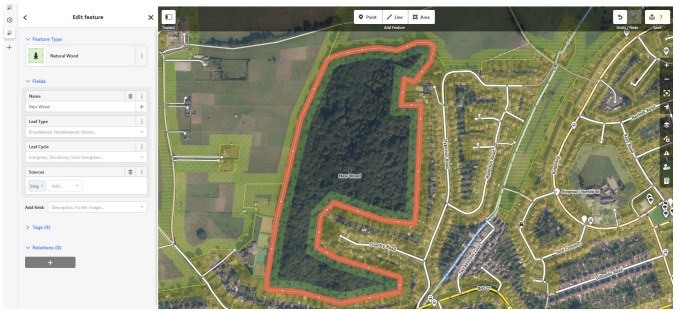

Fig. 11: The OpenStreetMap platform, which enables volunteers to use points, lines and areas to label ground objects on RSIs [165].

With the development of the Internet and information technology, user input interfaces, online labeling tools, and online storage tools have all been integrated into online services, making crowdsourcing more and more common in the field of RSI understanding. Social media (SM) network platforms (such as Twitter, Instagram, Facebook, Snapchat and Youtube) are increasingly used in data labeling [166], [167], [168]. Previous studies have shown that RSI and SM data can be combined and then used for disaster monitoring and environmental pollution monitoring, as well as identifying areas affected by disasters such as earthquakes, floods, and pollution leaks to help rescue [169], [44], [170], [171]. In addition, current smart devices can usually perform positioning, so they can collect social data with location information. Therefore, it is possible to combine crowdsourcing data and SM data for RSI labeling [172]. OpenStreetMap is a platform that combines crowdsourcing and social media to create a free editable map covering the world. In this platform, users can upload aerial images, manual survey data, GPS trajectory data and local information. Fig. 11 shows how to label the woodland on the OpenStreetMap platform.

With the increasing number of parameters of RSI understanding model, the demand for labeled data is increasing. In other words, it is difficult to train an excellent RSI understanding model based on deep neural network without enough labeled data [173], [174]. Therefore, in the field of RSI understanding, there are some researches using crowdsourcing to collect labeled data on RSIs, including using crowdsourcing to collect training data for RSI classification [175], [176], crop mapping [177], [178], human settlements mapping [179], [180], and crosswalk identification [181]. Therefore, one of the most important roles of crowdsourcing in the field of remote sensing is that it can support the training of various RSI understanding models and provide a fast, economic and reliable method to collect a large number of labeled samples. Crowdsourcing has great potential in solving the problem of limited labeled data in remote sensing. In the field of classical machine learning, many tasks (including image segmentation, image classification, and object detection [182]) use crowdsourcing method to collect labeled data, and have achieved remarkable results. However, the use of crowdsourcing in RSI understanding is still not common [183].

However, with the development of crowdsourcing platforms such as OpenStreetMap and Geo-Wiki, problems such as lack of labeled samples, difficulty in sample labeling, and high cost of sample labeling can be alleviated [184], [185]. It is easy to conclude that if a large number of labeled samples can be collected quickly and economically, the performance of various RSI understanding tasks may make great progress. By collecting enough training data from volunteers all over the world, the accuracy of land cover and land use classification model based on deep neural network will be improved, so as to quickly and accurately understand the global remote sensing data. Obtaining accurate global classification data will help to better understand various activities on the Earth, so as to obtain the evolution status of global forests, water bodies, crops and carbon resources. RSI understanding based on crowdsourcing is of great significance to further improve the depth and breadth of remote sensing applications [157].

With the continuous increase of remote sensing data obtained from various remote sensing platforms, it has become more and more difficult to obtain a sufficient amount of labeled data. When the various resolution data obtained from satellite platforms (such as Landsat, SPOT, MODIS, IKONOS, QuickBird and CBERS) and airborne platforms are gathered together, the amount of data will increase exponentially. With the current efficiency of manual labeling, it is difficult to keep up with the growth rate of remote sensing data. Therefore, crowdsourcing will become one of the important tools for understanding these remote sensing data in the future. It can be imagined that once the labeling problem of massive RSIs is solved through crowdsourcing, remote sensing researchers can create a powerful RSI understanding model, which can understand the RSIs collected on the remote sensing platform in real time and generate a time series. The time series will record the activities of the Earth in real time, making remote sensing one of the important tools to record the Earth and human history. Only by training the deep learning model with enough data can we reach this level in the field of RSI understanding.

## V. Conclusion

In recent years, RSI understanding based on supervised learning has achieved great success with a large number of noise-free training samples (i.e., strong supervised information). However, in practical remote sensing application tasks, it takes a lot of time and labor cost to collect supervised information. Therefore, researchers are exploring various methods for training the RSI understanding model with weakly supervised information, in this way, to improve the ability of remote sensing to serve the society.

This paper focused on three typical weakly supervised paradigms for RSI understanding: incomplete supervision, inexact supervision, and inaccurate supervision. Although these three paradigms have their own characteristics and applicable scenarios, in practice, they can also be used in RSI understanding at the same time, and there are some related researches on this hybrid weak supervision situation. For the future research of RSI understanding with weak supervision, we think it will be the combination of multiple weakly supervised paradigms, for example, the combination of inexact supervision and imprecise supervision, i.e., the given samples of RSI are not completely consistent with the strong supervised samples and contain noise. Another kind of hybrid weak supervision is the combination of incomplete supervision and inexact supervision, i.e., only a small number of the labeled samples can be obtained, and these labels are not fine-grained labels. In addition, the combination of self-supervision and weak supervision will also be a hot research direction in the future, because self-supervised learning can make full use of the self-supervised information of the RSI data itself to train a good RSI understanding model.

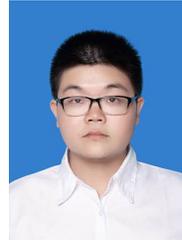

**Jun Yue** received the B.Eng. degree in geodesy from Wuhan University, Wuhan, China, in 2013 and the Ph.D. degree in GIS from Peking University, Beijing, China, in 2018.

He is currently an Assistant Professor with the Department of Geomatics Engineering, Changsha University of Science and Technology. His research interests include satellite image understanding, pattern recognition, and few-shot learning. Dr. Yue serves as a reviewer for IEEE Transactions on Geoscience and Remote Sensing, ISPRS Journal of Photogrammetry and Remote Sensing, IEEE Geoscience and Remote Sensing Letters, IEEE Transactions on Biomedical Engineering, IEEE Journal of Biomedical and health Informatics, Information Fusion, Information Sciences, International Journal of Remote Sensing, Remote Sensing Letters, etc.

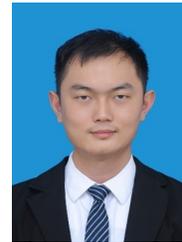

**Leyuan Fang** (Senior Member, IEEE) received the Ph.D. degree from the College of Electrical and Information Engineering, Hunan University, Changsha, China, in 2015.

From September 2011 to September 2012, he was a visiting Ph.D. student with the Department of Ophthalmology, Duke University, Durham, NC, USA, supported by the China Scholarship Council. From August 2016 to September 2017, he was a Postdoc Researcher with the Department of Biomedical Engineering, Duke University, Durham, NC, USA.

He is currently a Professor with the College of Electrical and Information Engineering, Hunan University, and an Adjunct Researcher with the Peng Cheng Laboratory. His research interests include sparse representation and multi-resolution analysis in remote sensing and medical image processing. He is the associate editors of IEEE Transactions on Image Processing, IEEE Transactions on Geoscience and Remote Sensing, IEEE Transactions on Neural Networks and Learning Systems, and Neurocomputing. He was a recipient of one 2nd-Grade National Award at the Nature and Science Progress of China in 2019.

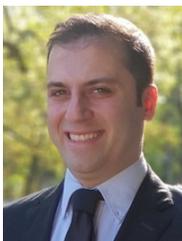

**Pedram Ghamisi** (Senior Member, IEEE) graduated with a Ph.D. in electrical and computer engineering at the University of Iceland in 2015. He works as (1) the head of the machine learning group at Helmholtz-Zentrum Dresden-Rossendorf (HZDR), Germany and (2) visiting professor and group leader of AI4RS at the Institute of Advanced Research in Artificial Intelligence (IARAI), Austria. He is a cofounder of VasoGnosis Inc. with two branches in San Jose and Milwaukee, the USA.

He was the co-chair of IEEE Image Analysis and Data Fusion Committee (IEEE IADF) between 2019 and 2021. Dr. Ghamisi was a recipient of the IEEE Mikio Takagi Prize for winning the Student Paper Competition at IEEE International Geoscience and Remote Sensing Symposium (IGARSS) in 2013, the first prize of the data fusion contest organized by the IEEE IADF in 2017, the Best Reviewer Prize of IEEE Geoscience and Remote Sensing Letters in 2017, and the IEEE Geoscience and Remote Sensing Society 2020 Highest-Impact Paper Award. His research interests include interdisciplinary research on machine (deep) learning, image and signal processing, and multisensor data fusion. He is also a co-founder of VasoGnosis Inc., with two branches in San, USA. For detailed info, please see http://pedram-ghamisi.com/.

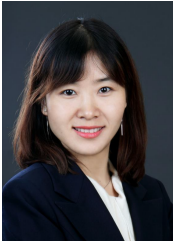

**Weiying Xie** (Member, IEEE) received the B.S. degree in electronic information science and technology from university of Jinan in 2011. She received the M.S. degree in communication and information systems, Lanzhou University in 2014 and the Ph.D. degree in communication and information systems of Xidian University in 2017. Currently, she is an Associate Professor with the State Key Laboratory of Integrated Services Networks, Xidian University. She has published more than 30 papers in refereed journals, including the IEEE TRANSACTIONS ON GEOSCIENCE AND REMOTE SENSING, the IEEE TRANSACTIONS ON NEURAL NETWORKS AND LEARNING SYSTEMS, the NEURAL NETWORKS, and the PATTERN RECOGNITION. Her research interests include neural networks, machine learning, hyperspectral image processing, and high-performance computing.

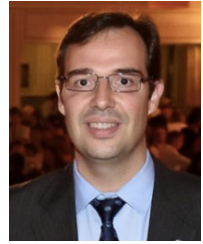

**Antonio Plaza** (Fellow, IEEE) received the M.Sc. and Ph.D. degrees in computer engineering from the University of Extremadura, Cáceres, Spain, in 1999 and 2002, respectively.

He is a Full Professor with the Department of Technology of Computers and Communications, University of Extremadura, where he is also the Head of the Hyperspectral Computing Laboratory (HyperComp). He has authored or coauthored more than 600 publications, including 324 journal citation report (JCR) articles (234 in IEEE journals), 25 international book chapters, and more than 300 peer-reviewed international conference papers. He has reviewed more than 500 manuscripts for more than 50 different journals. His main research interests include hyperspectral data processing and parallel computing of remote-sensing data.

Dr. Plaza is a Fellow of IEEE for his contributions to hyperspectral data processing and parallel computing of Earth observation data. He is a member of the Academy of Europe. He has served as the Editor-in-Chief for the IEEE TRANSACTIONS ON GEOSCIENCE AND REMOTE SENSING from 2013 to 2017. He is included in the Highly Cited Researchers List (Clarivate Analytics) from 2018 to 2020. (Additional information: http://www.umbc.edu/rssipl/people/aplaza.)

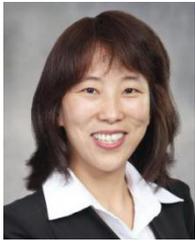

**Jun Li** (Fellow, IEEE) received the B.S. degree in geographic information systems from Hunan Normal University, Changsha, China, in 2004, the M.E. degree in remote sensing from Peking University, Beijing, China, in 2007, and the Ph.D. degree in electrical engineering from the Instituto de Telecomunicações, Instituto Superior Técnico (IST), Universidade Técnica de Lisbon, Lisbon, Portugal, in 2011.

She is a Full Professor with Sun Yat-sen University, Guangzhou, China. Her main research interests include remotely sensed hyperspectral image analysis, signal processing, active learning, and data fusion.

Dr. Li has received several important awards and distinctions, including the IEEE Geoscience and Remote Sensing Society (GRSS) Early Career Award in 2017, due to her outstanding contributions to remotely sensed hyperspectral and synthetic aperture radar data processing. She is the Editor-in-Chief of the IEEE JOURNAL OF SELECTED TOPICS IN APPLIED EARTH OBSERVATIONS AND REMOTE SENSING. She has been a Guest Editor of several journals, including the PROCEEDINGS OF THE IEEE and the ISPRS Journal of Photogrammetry and Remote Sensing.

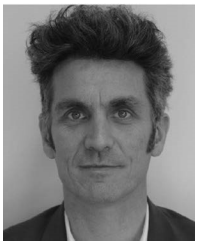

**Jocelyn Chanussot** (Fellow, IEEE) received the M.Sc. degree in electrical engineering from the Grenoble Institute of Technology (Grenoble INP), Grenoble, France, in 1995, and the Ph.D. degree from the University of Savoie, Annecy, France, in 1998.

Since 1999, he has been with Grenoble INP, where he was an Assistant Professor from 1999 to 2005 and an Associate Professor from 2005 to 2007 and is currently a Professor of signal and image processing. Since 2013, he has been an Adjunct Professor with the University of Iceland, Reykjavik, Iceland. His research interests include image analysis, multicomponent image processing, nonlinear filtering, and data fusion in remote sensing.

Dr. Chanussot was the founding President of the IEEE Geoscience and Remote Sensing French Chapter from 2007 to 2010, which is the recipient of the 2010 IEEE Geoscience and Remote Sensing Society (GRSS) Chapter Excellence Award. He was a co-recipient of the Nordic Signal Processing Symposium (NORSIG) 2006 Best Student Paper Award, the IEEE GRSS 2011 Symposium Best Paper Award, the IEEE GRSS 2012 Transactions Prize Paper Award, and the IEEE GRSS 2013 Highest Impact Paper Award.